\documentclass{article} 
\usepackage{arxiv/iclr2027_conference,times}

\usepackage{amsmath,amsfonts,bm}

\def\eqref#1{equation~\ref{#1}}

\def\1{\bm{1}}

\DeclareMathAlphabet{\mathsfit}{\encodingdefault}{\sfdefault}{m}{sl}
\SetMathAlphabet{\mathsfit}{bold}{\encodingdefault}{\sfdefault}{bx}{n}

\usepackage{url}
\usepackage{multirow} 
\usepackage{amsmath}
\usepackage{amssymb}
\usepackage{booktabs}
\usepackage{float}
\usepackage{graphicx}
\usepackage{xcolor}
\usepackage{placeins}
\usepackage{enumitem}
\setlist{nosep}

\newcommand{\blue}[1]{\textcolor{blue}{#1}}

\newcommand{\revise}[1]{\textcolor{blue}{#1}}

\newcommand{\parag}[1]{\vspace{-3mm}\paragraph{#1}}

\definecolor{ppblue}{HTML}{1F5FA8}   
\definecolor{ppred}{HTML}{B03A2E}    
\definecolor{ppurl}{HTML}{7A3E9D}    
\definecolor{oursrow}{HTML}{E8F1FB}  
\definecolor{ppgain}{HTML}{2E7D32}   
\definecolor{ppdrop}{HTML}{B03A2E}   
\definecolor{lightgray}{gray}{0.9}   
\definecolor{promptvar}{HTML}{008080}
\definecolor{pptoc}{HTML}{1F3A68}    

\usepackage{hyperref}
\hypersetup{
  colorlinks   = true,
  citecolor    = ppblue,
  linkcolor    = ppred,
  urlcolor     = ppurl,
  bookmarksnumbered = true,
  pdftitle     = {CogCanvas},
}
\title{Kingfisher: Fast Personalized Visual Autoregressive Generation via Scale-Aware Prompt-Conditioned Scale Decomposition}

\author{
Dinh-Khoi Vo$^{1,2}$ \quad
Akihiro Sugimoto$^{3,\dagger}$ \quad
Minh-Triet Tran$^{1,2}$ \quad
Trung-Nghia Le$^{1,2,\dagger,\ddagger}$ \\
$^{1}$University of Science, Ho Chi Minh City, Vietnam \\
$^{2}$Vietnam National University, Ho Chi Minh City, Vietnam\\
$^{3}$National Institute of Informatics, Tokyo, Japan \\
\texttt{vdkhoi@selab.hcmus.edu.vn} \quad
\texttt{sugimoto@nii.ac.jp} \\
\texttt{tmtriet@fit.hcmus.edu.vn} \quad
\texttt{ltnghia@fit.hcmus.edu.vn}\\
$^\dagger$Co-supervising.
$\ddagger$Corresponding author.\\
\textbf{Project page}:~\href{https://vdkhoi20.github.io/KINGFISHER}{{\setlength{\fboxsep}{1.5pt}\colorbox[HTML]{FDF0F4}{\strut\textcolor[HTML]{D6336C}{\texttt{https://vdkhoi20.github.io/KINGFISHER}}}}}.
}

\newcommand{\fix}{\marginpar{FIX}}
\newcommand{\new}{\marginpar{NEW}}

\iclrfinalcopy 
\begin{document}

\makeatletter
\g@addto@macro\@maketitle{
   \vspace{-8mm}
  \begin{figure}[H]
  \centering
  \includegraphics[width=\textwidth]{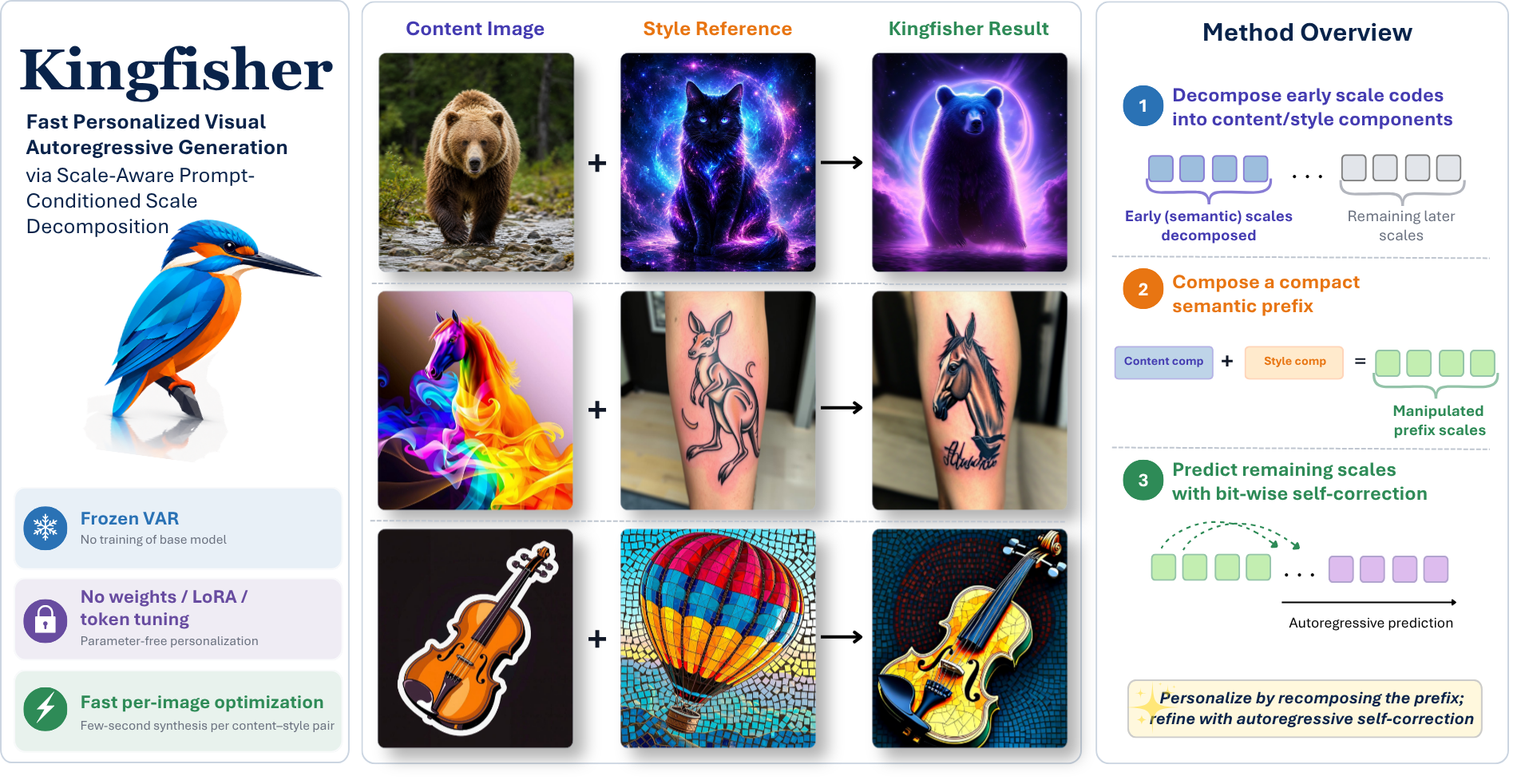}
  \vspace{-5mm}
	\caption{\textbf{Kingfisher} combines the content of one real image with the appearance of another by decomposing and recomposing early autoregressive scale codes. The remaining scales are predicted by the frozen VAR generator via bit-wise self-correction, enabling fast, coherent personalization without model tuning, LoRA, adapters, or learned subject embeddings.}
	\label{fig:teaser}
    \vspace{-2mm}
  \end{figure}
}
\makeatother

\maketitle

\begin{abstract}
Personalized image generation has been extensively studied in diffusion models, often relying on subject-specific optimization of model weights, textual embeddings, LoRA modules, or adapters. In this work, we explore a different direction for fast personalization in visual autoregressive models. Our key observation is that a short prefix of early autoregressive scale codes is sufficient to strongly determine both object structure and global appearance, while later scales primarily refine the established generation trajectory. Based on this insight, we introduce Kingfisher, a model-tuning-free framework that decomposes early residual scale codes into content- and style-related components through prompt-conditioned scale-code subspaces. Given two real images, Kingfisher first decomposes their early scale codes, then recomposes the content components of one image with the style components of the other into a compact semantic prefix. The frozen generator subsequently predicts the remaining scales through bitwise self-correction for coherent refinement. Experiments on CSD-100 show that Kingfisher achieves strong content preservation and the highest style and text alignment among the evaluated methods, while reducing pair-specific preparation time by approximately \(54\times\) compared with diffusion-based methods and \(7.5\times\) compared with VAR-based alternatives. These results demonstrate that effective personalization can be achieved by directly manipulating the native scale-space representation of a frozen visual autoregressive generator.

\end{abstract}

\section{Introduction}
Personalized image generation aims to adapt a generative model to user-provided visual concepts, enabling the synthesis of customized subjects, styles, or compositions under new conditions. Recent surveys~\cite{wei2025personalized,xu2025personalized} organize this problem across different generative architectures through inversion spaces~\cite{ho2020denoising}, inversion methods~\cite{song2020denoising}, and personalization schemes, covering GAN-based methods~\cite{richardson2021encoding,abdal2019image2stylegan}, diffusion personalization, and emerging autoregressive or multimodal generators. In diffusion models, existing methods personalize generation by operating in different representation spaces, including textual spaces~\cite{gal2022image,ruiz2023dreambooth}, feature spaces~\cite{wei2023elite,shi2024instantbooth}, noise spaces, and parameter spaces~\cite{hu2022lora,shah2024ziplora}. These methods are further realized through training-free inversion~\cite{mokady2023null}, optimization-based concept learning~\cite{kumari2023multi}, learning-based encoders or adapters~\cite{shi2024instantbooth,ye2023ip,ma2024subject}, and hybrid personalization pipelines~\cite{wang2024instantid,li2024photomaker}. Beyond single-subject customization, recent work also explores style-driven personalization and visual concept composition~\cite{vinker2023concept,frenkel2024implicit,dorfman2025ip,jin2024image,wang2025ms,gu2023mix}.

In parallel, visual autoregressive models have emerged as an alternative paradigm for image synthesis. In contrast to conventional token-by-token generation, Visual Autoregressive Modeling (VAR)~\cite{tian2024visual} formulates generation as coarse-to-fine next-scale prediction, while Infinity~\cite{han2025infinity} further introduces bitwise autoregressive modeling and self-correction for high-resolution text-to-image synthesis. This explicit scale-wise generation process suggests a different route for personalization: rather than adapting model parameters or auxiliary representations, one can directly operate on the native scale-code trajectory of a frozen generator. However, how content and style are organized across the scale hierarchy of visual autoregressive models remains underexplored. Diffusion-based personalization methods commonly represent visual factors through learned textual tokens, LoRA parameters, or image-conditioned features~\cite{gal2022image,ruiz2023dreambooth,frenkel2024implicit,liu2025unziplora}. Recent VAR-based works begin to reveal the importance of scale-wise representations: CSD-VAR~\cite{nguyen2025csd} studies scale-aware content-style decomposition, while training-free style personalization methods identify pivotal intermediate scales whose dominant components strongly affect style~\cite{lee2026training}. These works provide valuable evidence that individual scales encode different visual factors, but they largely analyze or intervene on scales independently. Motivated by this gap, through a series of controlled experiments, we examine consecutive scales as an autoregressive generation trajectory and observe that early scales act as semantic anchors that constrain subsequent visual refinements. This finding suggests that personalization need not modify the full generation sequence; composing a compact prefix is sufficient to steer content and appearance, while the frozen generator completes the remaining scales through autoregressive prediction.

We introduce \textbf{Kingfisher}, a model-tuning-free framework for fast personalized visual autoregressive generation. Motivated by the observation that early autoregressive scale codes are neither purely content nor purely style, but instead contain mixed content, appearance, and residual factors, Kingfisher models each scale code as an approximate composition \(R_i \approx C_i + S_i + e_i\), where \(C_i\) captures object-related structure, \(S_i\) captures visual appearance, and \(e_i\) absorbs residual information. To estimate these components without introducing new trainable parameters, Kingfisher constructs prompt-conditioned scale-code subspaces from the frozen generator itself: neutral prompts induce a neutral-style reference space, while object-specific prompts induce content directions relative to this reference. The decomposition is obtained by encouraging the style-removed code \(R_i-S_i\) to lie near the neutral-style subspace, the content-removed code \(R_i-C_i\) to suppress object-related directions, and the estimated content and style components to remain weakly coupled. Kingfisher then recomposes content factors from one real image and style factors from another within a compact semantic prefix, while the frozen autoregressive generator predicts the remaining scales through bitwise self-correction for coherent refinement. Although this formulation can be extended to other prompt-definable visual factors by constructing corresponding prompt-conditioned subspaces, we focus in this work on content--style personalization. We evaluate Kingfisher on CSD-100~\cite{nguyen2025csd} and show that, compared with prior diffusion-based personalization methods and VAR-based decomposition methods, it achieves a favorable trade-off between source-content preservation and reference-appearance adoption while substantially improving runtime efficiency. These results demonstrate effective content--style personalization without optimizing model weights, LoRA modules, adapters, or subject-specific embeddings. 

Our contributions are summarized as follows:
\begin{itemize}[leftmargin=2em]

\item We show that a short prefix of early autoregressive scale codes is sufficient to strongly control object structure and global appearance, revealing these scales as a compact semantic bottleneck for personalization.

\item We propose prompt-conditioned scale-code decomposition, which constructs neutral-style and object-content subspaces to explicitly decompose early scale codes into content- and style-related components in a frozen generator.

\item We introduce a scale-aware personalization pipeline that recomposes content and style components across two real images within a compact semantic prefix, while relying on autoregressive bitwise self-correction to generate the remaining scales for coherent refinement.

\item We demonstrate that Kingfisher achieves the highest style and text alignment among the evaluated methods on CSD-100 while maintaining strong content preservation, with approximately \(\mathbf{54\times}\) and \(\mathbf{7.5\times}\) faster pair-specific preparation than diffusion- and VAR-based alternatives, respectively.

\end{itemize}

\section{Related Work}

\parag{Personalized image generation.}
Personalized image generation has been extensively studied in diffusion models, where user-provided concepts are represented through textual embeddings~\cite{gal2022image}, subject-specific fine-tuning~\cite{ruiz2023dreambooth}, parameter-efficient updates such as LoRA~\cite{hu2022lora,shah2024ziplora,frenkel2024implicit}, or image-conditioning adapters and encoders~\cite{ye2023ip,wei2023elite,ma2024subject,shi2024instantbooth}. Recent methods further explore style-driven personalization, multi-concept composition, and concept disentanglement~\cite{vinker2023concept,dorfman2025ip,liu2025unziplora}. Beyond diffusion personalization, prior works on GAN inversion and latent editing also show that visual factors can be organized across generator layers or latent spaces. StyleGAN~\cite{karras2019style,karras2020training} provides scale-specific control through different synthesis layers, and StyleSpace further identifies localized channel-wise controls for distinct visual attributes~\cite{wu2021stylespace}. Inversion methods extend the latent space to \(\mathcal{W}+\)~\cite{tov2021designing} for more accurate reconstruction and editing of real images. Similar layer- or stage-wise control has been explored in diffusion models, where \(P+\)~\cite{voynov2023p+} uses per-layer prompt embeddings for more disentangled control and object--style mixing. These works suggest that generator layers or stages can expose controllable visual factors. In contrast, Kingfisher studies this question in visual autoregressive models and shows that early residual scales form entangled semantic bottlenecks, where object structure, appearance, composition, and style priors are coupled rather than cleanly separated.

\parag{Visual autoregressive generation.}
Autoregressive image generation models visual synthesis as sequential prediction~\cite{van2016conditional,reed2017parallel,chen2020generative}. Discrete image representation learning, such as VQ-VAE and VQGAN, provides quantized visual codes for compact image modeling~\cite{diederik2019introduction,van2017neural,razavi2019generating,esser2021taming}. When images are represented as discrete visual tokens, token-based autoregressive generators typically flatten them into one-dimensional code sequences and generate them in raster or predefined orders~\cite{esser2021taming,yu2022scaling}. In contrast, Visual Autoregressive Modeling (VAR) reformulates generation as coarse-to-fine next-scale prediction~\cite{tian2024visual}. Infinity further introduces bitwise visual autoregressive modeling and bitwise self-correction for high-resolution text-to-image synthesis~\cite{han2025infinity}. This explicit scale hierarchy motivates us to study how early scale codes can be decomposed and recomposed for fast personalization.

%

\begin{figure}[!t]
  \centering
  \includegraphics[width=0.8\textwidth]{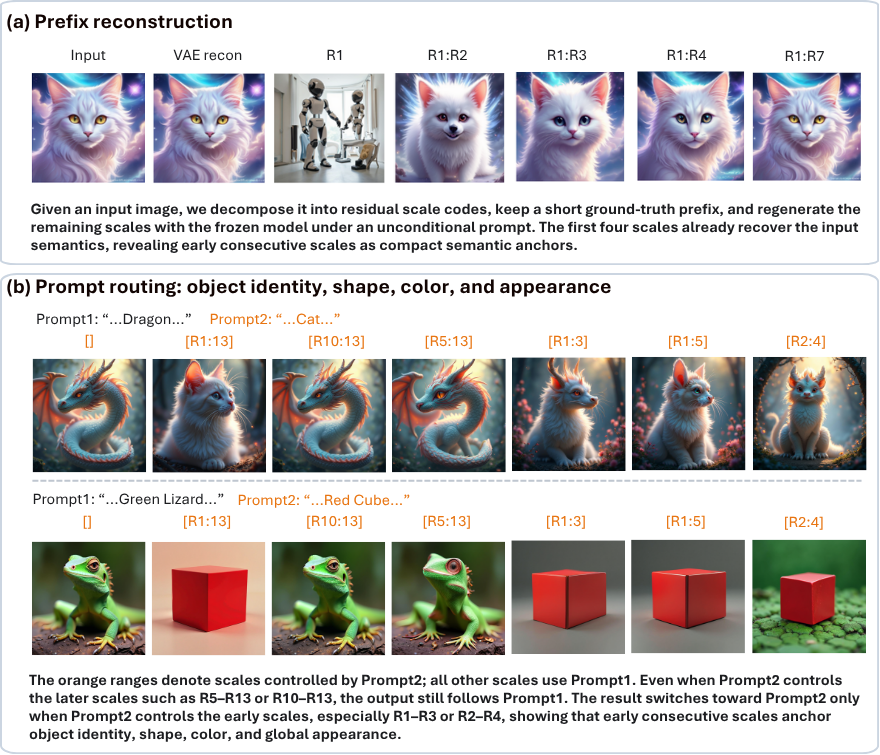}
  \vspace{-2mm}
  \caption{
\textbf{Scale-wise analysis of visual autoregressive generation.}
We analyze residual scales from two complementary views: prefix reconstruction and prompt routing. Prefix reconstruction shows that a short early ground-truth prefix is nearly sufficient to recover input semantics, while prompt routing shows that early scales dominate object identity, shape, color, and global appearance. Together, these results indicate that early consecutive scales form compact yet entangled semantic anchors, whereas later scales mainly refine details.  
}
    \vspace{-3mm}
  \label{fig:insights}
\end{figure}


\section{Scale-wise Insights and Preliminaries}
\label{sec:scale_insights}

\parag{Controlled scale-wise analysis.}
We analyze how visual information is organized across the residual scales of Infinity, where each image is represented by \(N=13\) residual scale codes \(\{R_i\}_{i=1}^{N}\). As shown in Fig.~\ref{fig:insights}, prefix reconstruction preserves the first \(k\) ground-truth scales and regenerates the remaining ones under an unconditional prompt; the first four scales already nearly reconstruct the input image. Prompt routing assigns Prompt$_2$ to selected scale ranges and Prompt$_1$ to the rest. Even when Prompt$_2$ controls large suffixes such as \(R_{5:13}\) or \(R_{10:13}\), the output still follows Prompt$_1$, and shifts toward Prompt$_2$ only when early scales such as \(R_{1:3}\) or \(R_{2:4}\) are routed to it. Appendix~\ref{app:scale_analysis} reports quantitative results for the prefix-reconstruction and prompt-routing experiments across different residual scale ranges. These results suggest that early residual scales act as semantic anchors, while later scales primarily refine local visual details.

\parag{Visual autoregressive generation and implication.}
VAR formulates image synthesis as coarse-to-fine next-scale prediction rather than raster-ordered token generation~\cite{tian2024visual}.We denote the residual scale codes by
\(\{R_i\}_{i=1}^{N}\), where \(N\) is the total number of scales. For compact notation, let
\(
R_{a:b} \triangleq (R_a,\ldots,R_b)\) and \(R_{<i} \triangleq (R_1,\ldots,R_{i-1}).
\)
The autoregressive generation process can then be factorized as:
\begin{equation}
p(R_{1:N}\mid y)=\prod_{i=1}^{N}p(R_i\mid R_{<i},y),
\end{equation}
where \(y\) is the text prompt. Infinity~\cite{han2025infinity} follows this scale-wise autoregressive formulation with bitwise visual token prediction and self-correction. At step \(i+1\), the model predicts \(R_{i+1}\) from the accumulated visual prefix:
\begin{equation}
Z_i=\sum_{j=1}^{i}\operatorname{Up}(R_j),
\end{equation}
where \(\operatorname{Up}(\cdot)\) interpolates each residual scale to the maximum spatial resolution \(H\times W\) before aggregation. The prediction also attends autoregressively to the previously computed prefix states \(\{Z_j\}_{j=1}^{i}\). Thus, predicting \(R_5\) depends on \(Z_4\), accumulated from \(R_{1:4}\), as well as earlier prefix states \(Z_{1:3}\). This repeated conditioning helps explain why early residual scales act as semantic anchors: later scales mostly refine details within the trajectory established by the early prefix. Therefore, the residual scale hierarchy is not a clean content-style separation; early consecutive scales instead form an entangled semantic bottleneck coupling object identity, shape, color, background, lighting, and style priors. This motivates our Kingfisher to decompose and recompose content and appearance factors only within the compact early prefix, while leaving the frozen generator to complete later scales through bitwise self-correction.

\begin{figure}[!t]
  \centering
  \includegraphics[width=\textwidth]{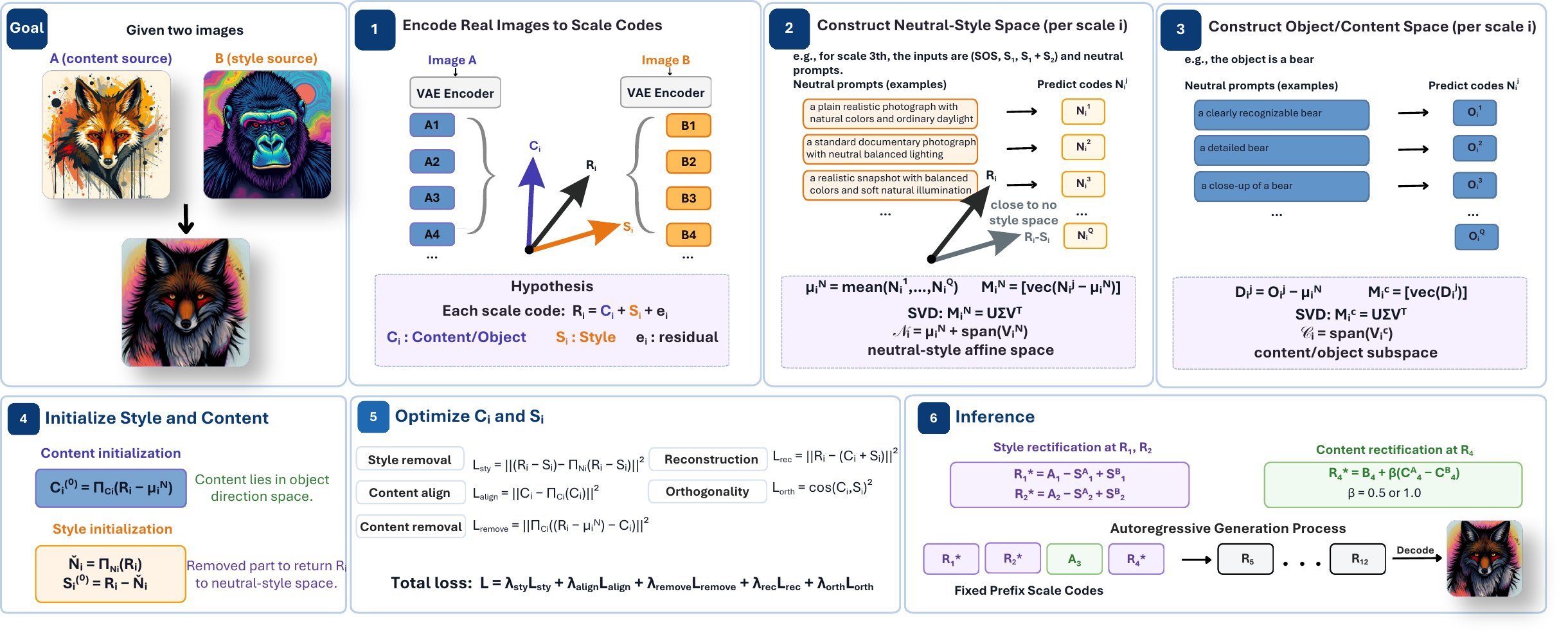}
  \vspace{-9mm}
  \caption{
  \textbf{Overview of the Kingfisher pipeline.}
  Given a content image \(A\) and a style image \(B\), both are encoded into multi-scale residual codes using the frozen tokenizer. Kingfisher constructs a neutral-style affine space from neutral prompts and an object-content subspace from object-specific prompts. The selected early residual codes are decomposed into content, style, and residual factors through prompt-space constraints and recomposed into a compact semantic prefix. The frozen Infinity generator then predicts the remaining scales through bitwise self-correction, producing an output that preserves the object structure of \(A\) while adopting the visual appearance of \(B\).
  }
  \vspace{-3mm}
  \label{fig:kingfisher_pipeline}
\end{figure}

\section{Proposed Method}
\label{sec:method}

\parag{Overview.}
As illustrated in Fig.~\ref{fig:kingfisher_pipeline}, given a content image \(I^A\) and a style reference \(I^B\), Kingfisher generates an image that preserves the object semantics and structure of \(I^A\) while adopting the visual appearance of \(I^B\). The frozen Infinity tokenizer encodes the two images into residual scale-code sequences \(\{A_i\}_{i=1}^{N}\) and \(\{B_i\}_{i=1}^{N}\), respectively. Guided by the scale-wise analysis in Sec.~\ref{sec:scale_insights}, Kingfisher operates only on the first four scales, where prompt-conditioned subspaces are used to estimate content and style components. These components are recomposed into a compact semantic prefix, while the frozen Infinity generator predicts the remaining scales through bitwise self-correction.

\subsection{Prompt-Conditioned Scale Subspaces}
\label{sec:subspaces}

For a real image encoded as \(\{R_i\}_{i=1}^{N}\), where N is number of residual scales, we query the frozen generator at scale \(i\)-th under prompt \(p\), while keeping the preceding scale-code prefix \(R_{<i}\) fixed:
\begin{equation}
    \widehat{R}_i(p)=\mathcal{G}_i(R_{<i},p).
\end{equation}

This isolates the influence of the prompt at scale \(i\) without changing its autoregressive image context.

Given neutral prompts \(\mathcal{P}^{N}=\{p_q^N\}_{q=1}^{Q_N}\), where the superscript \(N\) denotes neutral and \(Q_N\) is the number of neutral prompts, we obtain neutral predictions \(N_i^q=\widehat{R}_i(p_q^N)\). Their mean, centered matrix, and SVD are defined compactly as:
\begin{equation}
    \mu_i^N=\frac{1}{Q_N}\sum_{q=1}^{Q_N}N_i^q,\qquad
M_i^N=
\left[
\operatorname{vec}(N_i^q-\mu_i^N)^\top
\right]_{q=1}^{Q_N}
=
U_i^N\Sigma_i^N(V_i^N)^\top.
\end{equation}

Using the first \(r_N\) right singular vectors, we construct the neutral-style affine subspace:
\begin{equation}
    \mathcal{N}_i
=
\mu_i^N+\operatorname{span}(V_{i,1:r_N}^N).
\end{equation}

The neutral prompts describe natural colors, balanced illumination, and no explicit artistic stylization, providing an operational neutral-appearance reference.

Similarly, object prompts \(\mathcal{P}^{C}=\{p_q^C\}_{q=1}^{Q_C}\) produce \(O_i^q=\widehat{R}_i(p_q^C)\), where the superscript $C$ denotes content and $Q_C$ is number of content/object prompts. Since \(O_i^q\) is an absolute scale code, it contains both object-related information and shared appearance components. We therefore subtract the neutral-style mean \(\mu_i^N\) to remove this common baseline and isolate the code variation induced by the object prompt:
\begin{equation}
    D_i^q=O_i^q-\mu_i^N.
\end{equation}

Each \(D_i^q\) can thus be interpreted as a direction from the neutral reference toward the target object. We stack these directions, apply SVD, and define the object-content subspace as:
\begin{equation}
    M_i^C=
\left[\operatorname{vec}(D_i^q)^\top\right]_{q=1}^{Q_C}
=
U_i^C\Sigma_i^C(V_i^C)^\top,
\qquad
\mathcal{C}_i=\operatorname{span}(V_{i,1:r_C}^C).
\end{equation}

While \(\mathcal{N}_i\) models variations around a neutral appearance reference, sub space \(\mathcal{C}_i\) captures the principal directions associated with the target object.

\subsection{Prompt-Conditioned Scale Decomposition}
\label{sec:decomposition}

For each selected scale \(i\), we decompose the vectorized scale code as $R_i=C_i+S_i+e_i$, where \(C_i\), \(S_i\), and \(e_i\) denote object-content, visual-appearance, and unexplained residual components, respectively. 

\parag{Component initialization.}
Before joint optimization, we initialize \(C_i\) and \(S_i\) using the geometry of the two prompt-conditioned subspaces. Let
\(\bar{N}_i=\Pi_{\mathcal{N}_i}(R_i)\) be the closest neutral-style code to \(R_i\). 
We initialize:
\begin{equation}
    S_i^{(0)}=R_i-\bar{N}_i
\quad\Longrightarrow\quad
R_i-S_i^{(0)}=\bar{N}_i,
\qquad
C_i^{(0)}=\Pi_{\mathcal{C}_i}(R_i-\mu_i^N),
\end{equation}
where \(\Pi_{\mathcal{N}_i}(\cdot)\) and \(\Pi_{\mathcal{C}_i}(\cdot)\) denote projections onto the neutral-style and object-content subspaces, respectively.

Thus, \(S_i^{(0)}\) is exactly the component whose removal places \(R_i\) on the neutral-style subspace, while \(C_i^{(0)}\) extracts the object-related variation after removing the neutral baseline.
so removing the initial style component immediately yields a neutral-style code. For content, we first subtract the neutral mean to remove the shared appearance baseline, then project the remaining variation onto \(\mathcal{C}_i\). Thus, \(C_i^{(0)}\) is the object-aligned component of \(R_i\) that is best represented by the content subspace. These geometry-aware initializations provide meaningful starting points and reduce ambiguity during joint decomposition.

Starting from these initializations, we jointly optimize \(C_i\) and \(S_i\) using the following constraints:
\begin{equation}
\begin{alignedat}{2}
\mathcal{L}_{\mathrm{sty}}
&=\left\|(R_i-S_i)-\Pi_{\mathcal{N}_i}(R_i-S_i)\right\|_2^2,
&\qquad
\mathcal{L}_{\mathrm{align}}
&=\left\|C_i-\Pi_{\mathcal{C}_i}(C_i)\right\|_2^2,
\\
\mathcal{L}_{\mathrm{remove}}
&=\left\|\Pi_{\mathcal{C}_i}((R_i-\mu_i^N)-C_i)\right\|_2^2,
&
\mathcal{L}_{\mathrm{rec}}
&=\left\|R_i-(C_i+S_i)\right\|_2^2,
\\
&
\mathcal{L}_{\mathrm{orth}}=\cos^2(C_i,S_i).
&&
\end{alignedat}
\end{equation}

Here, \(\mathcal{L}_{\mathrm{sty}}\) makes the style-removed code close to the neutral subspace, while \(\mathcal{L}_{\mathrm{align}}\) constrains \(C_i\) to follow object-related directions. Since alignment alone does not ensure that \(C_i\) captures all content information, \(\mathcal{L}_{\mathrm{remove}}\) suppresses any content-subspace component remaining after \(C_i\) is removed. Finally, \(\mathcal{L}_{\mathrm{rec}}\) preserves the original information and \(\mathcal{L}_{\mathrm{orth}}\) reduces overlap between content and appearance. The total objective is:
\begin{equation}
    \mathcal{L}=
\lambda_{\mathrm{sty}}\mathcal{L}_{\mathrm{sty}}
+\lambda_{\mathrm{align}}\mathcal{L}_{\mathrm{align}}
+\lambda_{\mathrm{remove}}\mathcal{L}_{\mathrm{remove}}
+\lambda_{\mathrm{rec}}\mathcal{L}_{\mathrm{rec}}
+\lambda_{\mathrm{orth}}\mathcal{L}_{\mathrm{orth}}.
\end{equation}

\subsection{Scale-Aware Prefix Composition}
\label{sec:prefix_composition}
Given a content image \(A\) and a style image \(B\), we obtain \((C_i^A,S_i^A)\) and \((C_i^B,S_i^B)\) from the decomposition and recombine them to construct the early semantic prefix. Our preliminary experiments with different early-scale compositions show that the best balance between content preservation and appearance transfer is achieved by rectifying the appearance components at the first two scales, preserving the third scale as a semantic anchor, and correcting the fourth scale using a content direction. Specifically, we define:
\begin{equation}
R_i^\ast=A_i-S_i^A+S_i^B,\quad i\in\{1,2\},
\qquad
R_3^\ast=A_3,
\qquad
R_4^\ast=B_4+\beta(C_4^A-C_4^B),
\end{equation}
where scale code \(A\) provides the target content, scale code \(B\) provides the reference appearance, and \(\beta\) controls the strength of the structural correction at \(R_4\). The resulting semantic prefix is:
\begin{equation}
\mathcal{P}^\ast=[R_1^\ast,R_2^\ast,A_3,R_4^\ast].
\end{equation}

At \(R_1\) and \(R_2\), appearance rectification retains the original content code from \(A\) while replacing only its estimated appearance component with that of \(B\). In contrast, \(A_3\) is preserved directly because modifying this scale tends to disturb object identity and structure. Finally, \(R_4^\ast\) uses \(B_4\) as an appearance-aware base and adds the content difference \(C_4^A-C_4^B\) to steer it toward the structure of \(A\). Thus, the prefix combines coarse appearance transfer, a preserved semantic anchor, and content-aware structural correction.

We do not copy the remaining scales from either image, since later real-image codes may reintroduce source-specific details. Instead, the frozen generator autoregressively predicts the suffix conditioned on the composed prefix and target prompt \(y^\ast\):
\begin{equation}
p(R_{5:N}^\ast\mid\mathcal{P}^\ast,y^\ast)
=
\prod_{i=5}^{N}
p_\theta(R_i^\ast\mid R_{<i}^\ast,y^\ast),
\qquad \theta\ \text{frozen}.
\end{equation}

The generated image is decoded as:
\begin{equation}
I^\ast
=
\operatorname{Dec}
\left(
\sum_{i=1}^{N}\operatorname{Up}(R_i^\ast)
\right).
\end{equation}

Infinity's bitwise self-correction completes the remaining scales while maintaining consistency with the composed semantic prefix. Although we focus on content--style personalization, the same formulation can support other prompt-definable factors by constructing their corresponding scale-code subspaces.

\section{Experiments}
\label{sec:experiments}

\subsection{Experimental Setup}

\parag{Dataset and evaluation protocol.}
We evaluate Kingfisher on the CSD-100 dataset~\cite{nguyen2025csd}, which contains 100 real images spanning diverse object and appearance concepts. Since our task takes two real images as input, we construct ordered cross-image pairs \((I^A,I^B)\), where \(I^A\) is the content image
and \(I^B\) is the style reference image. For each content image, we sample five style references from different categories while excluding pairs sharing the same object category. This results in 500 ordered content--style pairs. We generate one output for each pair, yielding 500 images per method. For each pair, the target semantic prompt is defined as ``a photo of \texttt{<content-A>} in \texttt{<style-B>} style.''
We use a fixed seed of 2026 for reproducibility and do not perform best-of-seed selection.

\parag{Baselines.} 
We compare Kingfisher with representative personalization and content--style composition methods from both diffusion and visual autoregressive families: DreamBooth-LoRA~\cite{ruiz2023dreambooth}, B-LoRA~\cite{frenkel2024implicit}, UnZipLoRA~\cite{liu2025unziplora}, Inspiration Tree~\cite{vinker2023concept}, and CSD-VAR~\cite{nguyen2025csd}. We use publicly available implementations released by the original authors whenever available, and explicitly document any adaptation required for our cross-image setting or backbone compatibility. The diffusion baselines are implemented on SDXL~\cite{podell2024sdxl}.


\parag{Metrics.} We evaluate content preservation, style alignment, and text alignment using five metrics. \textit{CSD-C} and \textit{CSD-S} are computed using the content and style embeddings extracted by the CSD evaluator~\cite{nguyen2025csd}. Specifically, CSD-C is the cosine similarity between the content embedding of the generated image and that of the content image \(I^A\), whereas CSD-S is the cosine similarity between the style embedding of the generated image and that of the style image \(I^B\). \textit{CLIP-I} measures the cosine similarity between the CLIP image features~\cite{radford2021learning} of the generated image and \(I^A\). \textit{DINO} measures the cosine similarity between DINO visual features~\cite{oquab2023dinov2} of the generated image and \(I^B\). Finally, \textit{CLIP-T} measures the cosine similarity between the CLIP image feature of the generated image and the CLIP text feature of the target content--style prompt. Higher values indicate better performance for all metrics.



\subsection{Comparison with Prior Methods}

\parag{Quantitative comparison.} 

\begin{table}[!t]
\centering
\caption{Quantitative comparison on the cross-image CSD-100 benchmark.
Best results are shown in \textbf{bold} and second-best results are \underline{underlined}.}
\label{tab:main_comparison}
\scriptsize
\setlength{\tabcolsep}{3.2pt}
\renewcommand{\arraystretch}{1.05}
\begin{tabular}{lccccc}
\toprule
& \multicolumn{2}{c}{\textbf{Content Align.}}
& \multicolumn{2}{c}{\textbf{Style Align.}}
& \textbf{Text Align.} \\
\cmidrule(lr){2-3}\cmidrule(lr){4-5}
\textbf{Method}
& \textbf{CSD-C} $\uparrow$
& \textbf{CLIP-I} $\uparrow$
& \textbf{CSD-S} $\uparrow$
& \textbf{DINO} $\uparrow$
& \textbf{CLIP-T} $\uparrow$ \\
\midrule

DreamBooth-LoRA~\cite{ruiz2023dreambooth}
& \underline{0.5519}
& 0.7362
& 0.4207
& 0.3813
& 0.2492 \\

B-LoRA~\cite{frenkel2024implicit}
& 0.4739
& 0.7182
& \underline{0.4990}
& \underline{0.4220}
& \underline{0.2683} \\

UnZipLoRA~\cite{liu2025unziplora}
& \textbf{0.5782}
& 0.7322
& 0.3673
& 0.3560
& 0.2472 \\

Inspiration Tree~\cite{vinker2023concept}
& 0.5413
& \underline{0.7591}
& 0.2897
& 0.2601
& 0.2565 \\

CSD-VAR~\cite{nguyen2025csd}
& 0.4731
& 0.7184
& 0.2812
& 0.2503
& 0.2648 \\


\midrule
\textbf{Kingfisher (Ours)}
& \underline{0.5519}
& \textbf{0.7900}
& \textbf{0.5222}
& \textbf{0.4350}
& \textbf{0.3105} \\

\bottomrule
\end{tabular}
\vspace{-3mm}
\end{table}

Table~\ref{tab:main_comparison} reports quantitative results over all 500 ordered content--style pairs. Kingfisher achieves the best performance on CLIP-I, CSD-S, DINO, and CLIP-T, showing a strong balance between content preservation and appearance transfer. DreamBooth-LoRA preserves content reasonably well but shows weaker style alignment, which is consistent with the residual entanglement between content and appearance in parameter-efficient personalization. B-LoRA and UnZipLoRA explicitly separate content- and style-related representations and therefore achieve stronger cross-image composition, with B-LoRA favoring style transfer and UnZipLoRA favoring content preservation. Inspiration Tree retains content well but transfers style less effectively in our setting. Among visual autoregressive methods, CSD-VAR explicitly decomposes content and style representations, yet Kingfisher performs substantially better. This suggests that directly manipulating the early residual scale codes provides a more effective way to control the content--style decomposition.

\parag{Qualitative comparison.} 

\begin{figure}[t]
\centering
\includegraphics[width=0.75\textwidth]{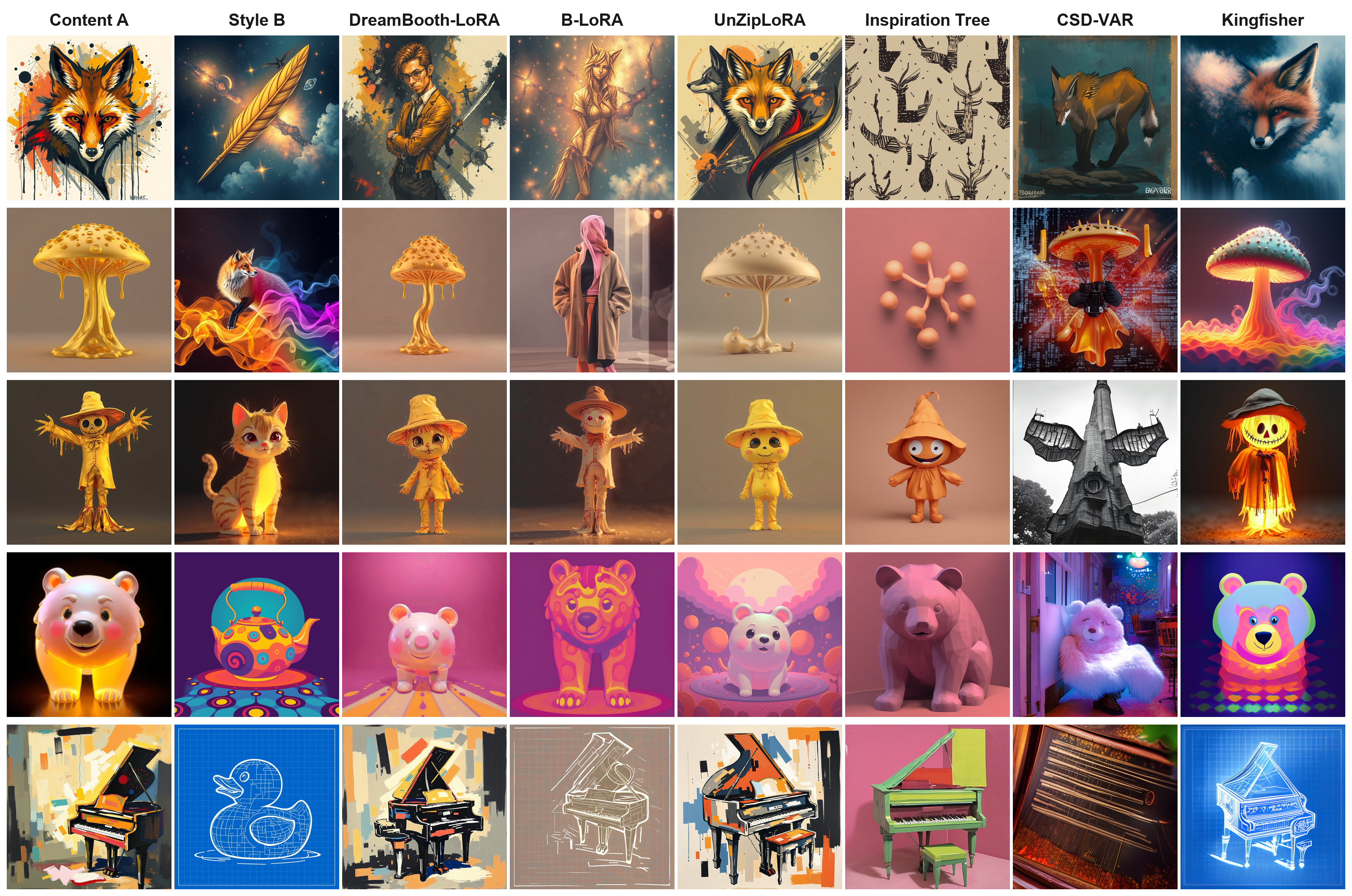}
\vspace{-4mm}
\caption{Qualitative comparison on the cross-image CSD-100 benchmark~\cite{nguyen2025csd}. Given a content reference \(I^A\) and a style reference \(I^B\), each method aims to preserve the content of \(I^A\) while transferring the appearance of \(I^B\). Kingfisher achieves a favorable balance between content preservation and style transfer across diverse content--style combinations.}
\vspace{-5mm}
\label{fig:qualitative_comparison}
\end{figure}

Figure~\ref{fig:qualitative_comparison} compares representative cross-image compositions under identical content--style pairs. Existing personalization methods typically obtain content and style through learned implicit representations, such as tokens, embeddings, or parameter updates. Since these representations are not explicitly constrained to encode a single factor, content and appearance may remain entangled. As a result, implicit personalization often benefits from multiple reference images of the same object under different backgrounds or conditions to isolate a more consistent object representation, and similarly from multiple examples sharing the same visual style to suppress content-specific cues. With only a single reference image, the learned representation can easily absorb irrelevant appearance or structural information, leading to overfitting to one reference or unnatural compositions when content and style representations are combined. In contrast, Kingfisher explicitly estimates content- and appearance-related components in the early residual scale codes and recomposes them before autoregressive refinement. This leads to more coherent outputs that preserve the object and spatial structure of \(I^A\) while transferring the visual appearance of \(I^B\).

\parag{Efficiency Analysis} 

\begin{table}[!t]
\centering
\caption{Efficiency and human evaluation.
Efficiency is measured on a single A100 GPU. Preparation denotes the pair-specific optimization time for one content--style pair, while inference denotes the time required to generate one image after preparation. Human preference is computed from 600 forced-choice selections.}
\label{tab:efficiency_human}
\scriptsize
\setlength{\tabcolsep}{3pt}
\renewcommand{\arraystretch}{1.08}
\begin{tabular}{lcccc}
\toprule
& \multicolumn{2}{c}{\textbf{Efficiency}}
& \multicolumn{2}{c}{\textbf{Human Evaluation}} \\
\cmidrule(lr){2-3}
\cmidrule(lr){4-5}
\textbf{Method}
& \textbf{Prep./pair} (s) $\downarrow$
& \textbf{Infer./image} (s) $\downarrow$
& \textbf{Votes} $\uparrow$
& \textbf{Pref.} $\uparrow$ \\
\midrule
DreamBooth-LoRA~\cite{ruiz2023dreambooth}
& $1542.21 \pm 38.47$
& $11.98 \pm 0.91$
& 54
& 9.0\% \\

B-LoRA~\cite{frenkel2024implicit}
& $1309.00 \pm 27.63$
& $12.88 \pm 1.02$
& \underline{168}
& \underline{28.0\%} \\

UnZipLoRA~\cite{liu2025unziplora}
& $3463.70 \pm 40.28$
& $11.83 \pm 0.95$
& 45
& 7.5\% \\

Inspiration Tree~\cite{vinker2023concept}
& $7465.36 \pm 216.84$
& $6.56 \pm 0.54$
& 39
& 6.5\% \\

CSD-VAR~\cite{nguyen2025csd}
& $\underline{171.65 \pm 11.82}$
& $\mathbf{1.39 \pm 0.12}$
& 40
& 6.7\% \\

\midrule
\textbf{Kingfisher (Ours)}
& $\mathbf{23.54 \pm 1.86}$
& $\underline{1.61 \pm 0.15}$
& \textbf{254}
& \textbf{42.3\%} \\
\bottomrule
\end{tabular}
\vspace{-6mm}
\end{table}

Table~\ref{tab:efficiency_human} compares the per-pair computational cost. Unlike methods that optimize model-side weights, LoRA adapters, or learned subject representations, Kingfisher operates directly on the early residual scale codes while keeping the complete Infinity generator frozen. Kingfisher requires only \(23.54\) seconds of preparation per pair, compared with \(171.65\) seconds for CSD-VAR. This corresponds to more than a \(10\times\) reduction in preparation cost. Kingfisher also requires only \(1.61\) seconds to generate an output, remaining close to the fastest autoregressive baseline while being substantially faster than the diffusion-based methods.







\parag{Human Evaluation}
\label{sec:human_evaluation}

We conducted a blind forced-choice study with 20 participants on 600 responses to compare
Kingfisher with existing five methods. For each pair of content-style,
participants selected the output that best preserved the content image,
transferred the style reference, and maintained overall image quality. As shown in Table~\ref{tab:efficiency_human}, Kingfisher receives the highest human preference among the evaluated methods, followed by B-LoRA. The results indicate that participants more frequently prefer Kingfisher when jointly considering content preservation, appearance transfer, and overall visual quality, consistent with the trends observed in our qualitative and automatic evaluations.

\section{Discussion}
\label{sec:discussion}

Our results highlight a different perspective on personalization in visual autoregressive models. Many existing personalization approaches represent content and style implicitly through learned tokens, embeddings, adapters, or parameter updates, where the two factors may remain entangled within the learned representation. In contrast, Kingfisher explicitly models each early residual scale in terms of content- and appearance-related components and directly constrains their roles through prompt-conditioned scale subspaces. This formulation provides an explicit mechanism for analyzing and recomposing content and appearance within the native residual scale trajectory of a frozen generator. Our scale-wise analysis further shows that these factors are not naturally separated across individual scales; instead, the early scales form an entangled semantic bottleneck that jointly determines object structure and global appearance. By operating only within this compact early prefix, Kingfisher also substantially reduces the personalization cost compared with optimization-heavy alternatives.
\section{Conclusion}

We introduced Kingfisher, a fast personalization framework for visual autoregressive models that operates directly on the residual scale trajectory of a frozen pretrained generator. By explicitly decomposing and recomposing content- and appearance-related components within the early semantic prefix, Kingfisher achieves effective content--style personalization without model fine-tuning. Experiments on CSD-100 dataset show that Kingfisher achieves strong content–style composition and substantially reduces personalization cost. These results demonstrate that native scale trajectories can provide an efficient interface for personalized visual autoregressive generation without model fine-tuning.



\section*{AI Use Statement}

ChatGPT was used solely for grammar correction and language polishing. All technical methodology, experiments, and results were developed, conducted, and verified by the authors.

\section*{Ethics Statement}
Our human evaluation involves voluntary and anonymous preference judgments on generated images. No personal identifiers or sensitive information were collected, and the responses were used solely to evaluate perceptual preferences among the compared methods. 

\section*{Reproducibility Statement}
We provide detailed descriptions of the CSD-100 evaluation protocol, baseline configurations, evaluation metrics, implementation settings, and hyperparameters in Sec.~\ref{sec:experiments}. Additional qualitative results, ablation studies, and the exact neutral and object-conditioned prompt templates used to construct the prompt-conditioned subspaces are provided in the appendix. These details are intended to facilitate reproduction of the reported results.

\bibliography{iclr2027_conference}
\bibliographystyle{iclr2027_conference}

\FloatBarrier

\clearpage
\appendix

\section*{Appendix}

\section{Additional Scale-wise Analysis}
\label{app:scale_analysis}

Table~\ref{tab:scale_role_analysis} reports the quantitative results of the controlled prefix-reconstruction and prompt-routing experiments described in Sec.~\ref{sec:scale_insights}.

\begin{table}[H]
\centering
\scriptsize
\setlength{\tabcolsep}{3pt}
\renewcommand{\arraystretch}{0.92}
\caption{\textbf{Quantitative scale-wise analysis.} Results are averaged over 100 real images and 200 prompt-routing pairs. For prefix reconstruction, CLIP-I and DINO-I measure similarity to the input image, while LPIPS measures perceptual reconstruction distance. For prompt routing, CLIP-T measures alignment with Prompt$_2$, while DINO-P1 measures similarity to the all-Prompt$_1$ reference.}
\label{tab:scale_role_analysis}
\begin{tabular}{lcccccc}
\toprule
Experiment & Range
& CLIP-I $\uparrow$
& DINO-I $\uparrow$
& LPIPS $\downarrow$
& CLIP-T $\uparrow$
& DINO-P1 $\uparrow$ \\
\midrule
Prefix recon. & \(R_1\)     & 0.42 & 0.36 & 0.58 & -- & -- \\
Prefix recon. & \(R_{1:2}\) & 0.61 & 0.54 & 0.41 & -- & -- \\
Prefix recon. & \(R_{1:3}\) & 0.78 & 0.72 & 0.26 & -- & -- \\
Prefix recon. & \(R_{1:4}\) & 0.88 & 0.84 & 0.16 & -- & -- \\
Prefix recon. & \(R_{1:7}\) & 0.93 & 0.89 & 0.11 & -- & -- \\
\midrule
Prompt routing & \(R_{10:13}\) & -- & -- & -- & 0.28 & 0.74 \\
Prompt routing & \(R_{5:13}\)  & -- & -- & -- & 0.34 & 0.70 \\
Prompt routing & \(R_{1:3}\)   & -- & -- & -- & 0.76 & 0.39 \\
Prompt routing & \(R_{2:4}\)   & -- & -- & -- & 0.81 & 0.35 \\
Prompt routing & \(R_{1:5}\)   & -- & -- & -- & 0.84 & 0.31 \\
\bottomrule
\end{tabular}
\end{table}

\section{Condition-Prompt Construction}
\label{app:prompt_construction}

Kingfisher constructs the prompt-conditioned scale subspaces using two prompt sets. The neutral prompts describe realistic images with ordinary colors, illumination, texture, and background, providing a neutral appearance reference. The object-conditioned prompts describe the target object through multiple semantic formulations, providing diverse object-related directions while reducing sensitivity to a particular wording. Table~\ref{tab:condition_prompts} lists the prompt templates used in our experiments, where \texttt{<object>} is replaced by the object category of each input image.

\begin{table}[H]
\centering
\caption{Neutral and object-conditioned prompt templates used to construct the prompt-conditioned scale subspaces.}
\label{tab:condition_prompts}
\scriptsize
\setlength{\tabcolsep}{5pt}
\renewcommand{\arraystretch}{1.05}
\begin{tabular}{cl}
\toprule
Type & Prompt template \\
\midrule
\multirow{5}{*}{Neutral}
& ``a plain realistic photograph with natural colors and ordinary daylight'' \\
& ``a standard documentary photograph with neutral balanced lighting'' \\
& ``a clean product photograph with a simple background and natural texture'' \\
& ``a realistic snapshot with balanced colors and soft natural illumination'' \\
& ``a straightforward studio photograph with ordinary materials and neutral tones'' \\
\midrule
\multirow{6}{*}{Object}
& ``a clearly recognizable \texttt{<object>}'' \\
& ``a detailed \texttt{<object>}'' \\
& ``a centered \texttt{<object>}'' \\
& ``a close-up view of a \texttt{<object>}'' \\
& ``a \texttt{<object>} with a clear silhouette'' \\
& ``a realistic photograph of a \texttt{<object>}'' \\
\bottomrule
\end{tabular}
\end{table}

The neutral predictions are used to construct the neutral-style affine subspace \(\mathcal{N}_i\), while the object-conditioned predictions are centered relative to the neutral reference and used to construct the object-content subspace \(\mathcal{C}_i\). These prompts are used only to query the frozen generator and introduce no learned textual embeddings or additional trainable parameters.

\section{Details of Human Evaluation}
 
We conduct a human preference study to complement the automatic evaluation by presenting the generated results from all evaluated methods side by side. Each question presents one content--style pair consisting of the content reference \(I^A\), the style reference \(I^B\), and six anonymous outputs generated by Kingfisher, DreamBooth-LoRA~\cite{ruiz2023dreambooth}, B-LoRA~\cite{frenkel2024implicit}, UnZipLoRA~\cite{liu2025unziplora}, Inspiration Tree~\cite{vinker2023concept}, and CSD-VAR~\cite{nguyen2025csd}. Participants are asked to select the single output that best jointly preserves the object identity and spatial structure of \(I^A\), adopts the visual appearance of \(I^B\), and maintains overall image quality. Method identities are hidden, and both the order of the content--style pairs and the order of the seven method outputs within each question are independently randomized for each questionnaire version. To reduce fatigue, the study is divided into 10 questionnaire versions, each containing 30 unique pairs, covering 300 of the 500 benchmark pairs in total. We collected responses from 20 participants, resulting in 600 total selections. Depending on the assigned questionnaire version, each evaluated content--style pair receives between one and three independent evaluations. We aggregate all valid selections and report the percentage of preferences received by each method.

\section{Ablation Studies}

\paragraph{Early-scale prefix composition.} We investigate how progressively rectifying the early residual scales affects the content--style trade-off. Table~\ref{tab:prefix_ablation} shows that increasing the number of rectified scales consistently improves style alignment while reducing content preservation. The original prefix \([A_1,A_2,A_3,A_4]\) achieves the strongest content scores but the weakest style scores. Introducing \(R_1^\ast\) and \(R_4^\ast\) improves style alignment, and further rectifying \(R_2\) provides a more balanced trade-off between content and style while achieving the highest CLIP-T score. Rectifying \(R_3\) as well yields the strongest style alignment but at the cost of substantially lower content similarity. We therefore adopt \([R_1^\ast,R_2^\ast,A_3,R_4^\ast]\) as the default configuration to maintain a balanced trade-off between content preservation and appearance transfer. 

\begin{table}[H]
\centering
\caption{Ablation of early-scale prefix composition. Best results are shown in bold and second-best results are underlined.}
\label{tab:prefix_ablation}
\scriptsize
\setlength{\tabcolsep}{2.8pt}
\renewcommand{\arraystretch}{1.05}
\begin{tabular}{lccccc}
\toprule
Prefix configuration
& CSD-C $\uparrow$
& CLIP-I $\uparrow$
& CSD-S $\uparrow$
& DINO $\uparrow$
& CLIP-T $\uparrow$ \\
\midrule
\([A_1,A_2,A_3,A_4]\)
& \textbf{0.7774}
& \textbf{0.9370}
& 0.2922
& 0.2720
& 0.3017 \\
\([R_1^\ast,A_2,A_3,R_4^\ast]\)
& \underline{0.6281}
& \underline{0.8392}
& 0.4342
& 0.3751
& \underline{0.3119} \\
\(\mathbf{[R_1^\ast,R_2^\ast,A_3,R_4^\ast]}\)
& 0.5504
& 0.7922
& \underline{0.5210}
& \underline{0.4367}
& \textbf{0.3127} \\
\([R_1^\ast,R_2^\ast,R_3^\ast,R_4^\ast]\)
& 0.4979
& 0.7416
& \textbf{0.6156}
& \textbf{0.5416}
& 0.2959 \\
\bottomrule
\end{tabular}
\end{table}

\paragraph{Structural correction strength.} We next study the coefficient \(\beta\), which controls the contribution of the content correction at the fourth residual scale, \(R_4^\ast=B_4+\beta(C_4^A-C_4^B)\). Table~\ref{tab:beta_ablation} reveals a clear trade-off: as \(\beta\) increases, CSD-C and CLIP-I consistently improve, whereas CSD-S and DINO decrease. Larger values therefore favor content preservation, while smaller values retain stronger appearance alignment. We use \(\beta=0.5\) as an intermediate operating point that balances these two tendencies without strongly favoring either content or style.

\begin{table}[H]
\centering
\caption{Sensitivity to the structural correction coefficient \(\beta\). Best results are shown in bold and second-best results are underlined.}
\label{tab:beta_ablation}
\scriptsize
\setlength{\tabcolsep}{3.5pt}
\renewcommand{\arraystretch}{1.05}
\begin{tabular}{cccccc}
\toprule
\(\beta\)
& CSD-C $\uparrow$
& CLIP-I $\uparrow$
& CSD-S $\uparrow$
& DINO $\uparrow$
& CLIP-T $\uparrow$ \\
\midrule
0.00
& 0.5436
& 0.7784
& \textbf{0.5353}
& \textbf{0.4597}
& 0.3068 \\
0.25
& 0.5471
& 0.7866
& \underline{0.5333}
& \underline{0.4521}
& 0.3104 \\
\textbf{0.50}
& 0.5504
& 0.7922
& 0.5210
& 0.4367
& 0.3127 \\
0.75
& \underline{0.5551}
& \underline{0.7968}
& 0.5161
& 0.4239
& \underline{0.3130} \\
1.00
& \textbf{0.5613}
& \textbf{0.8027}
& 0.5063
& 0.4096
& \textbf{0.3144} \\
\bottomrule
\end{tabular}
\end{table}

\paragraph{Decomposition constraints.} Table~\ref{tab:loss_ablation} evaluates the contribution of each term in the decomposition objective. Removing \(\mathcal{L}_{\mathrm{sty}}\) mainly degrades style alignment, confirming its role in isolating the appearance component. Removing the content alignment or removal constraints primarily reduces content preservation, as the estimated content component is no longer sufficiently constrained to represent and isolate object-related information. Without \(\mathcal{L}_{\mathrm{rec}}\), both content and style performance decrease substantially because information is no longer preserved during decomposition. Finally, removing \(\mathcal{L}_{\mathrm{orth}}\) causes the content and style components to overlap, leading to cross-factor leakage and moderate degradation in both aspects. Overall, the full objective provides the most consistent balance across content preservation, style transfer, and text alignment.

\begin{table}[H]
\centering
\caption{Ablation of the decomposition objective. Best results are shown in bold and second-best results are underlined.}
\label{tab:loss_ablation}
\scriptsize
\setlength{\tabcolsep}{2.8pt}
\renewcommand{\arraystretch}{1.05}
\begin{tabular}{lccccc}
\toprule
Configuration
& CSD-C $\uparrow$
& CLIP-I $\uparrow$
& CSD-S $\uparrow$
& DINO $\uparrow$
& CLIP-T $\uparrow$ \\
\midrule
w/o \(\mathcal{L}_{\mathrm{sty}}\)
& \underline{0.5478}
& \underline{0.7895}
& 0.4726
& 0.3914
& 0.3048 \\

w/o \(\mathcal{L}_{\mathrm{align}}\)
& 0.5127
& 0.7568
& \underline{0.5073}
& \underline{0.4218}
& 0.3035 \\

w/o \(\mathcal{L}_{\mathrm{remove}}\)
& 0.5019
& 0.7446
& 0.4981
& 0.4127
& 0.3009 \\

w/o \(\mathcal{L}_{\mathrm{rec}}\)
& 0.4742
& 0.7165
& 0.4386
& 0.3597
& 0.2863 \\

w/o \(\mathcal{L}_{\mathrm{orth}}\)
& 0.5264
& 0.7698
& 0.4935
& 0.4062
& \underline{0.3074} \\
\midrule
\textbf{Full objective}
& \textbf{0.5504}
& \textbf{0.7922}
& \textbf{0.5210}
& \textbf{0.4367}
& \textbf{0.3127} \\
\bottomrule
\end{tabular}
\end{table}



\begin{figure}[H]
\centering
\includegraphics[width=\textwidth]{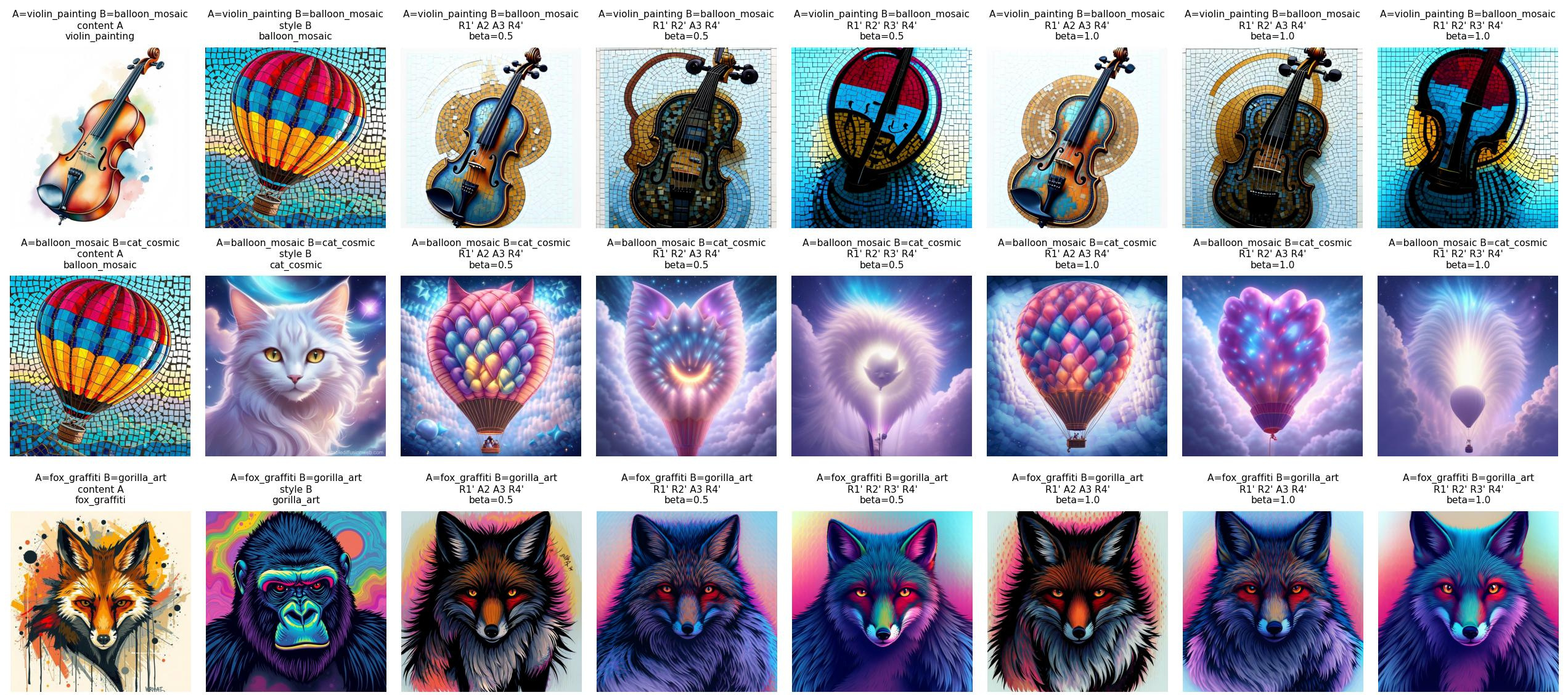}
\caption{\textbf{Additional qualitative ablation of prefix composition.} From left to right: content reference \(I^A\), style reference \(I^B\), \([R_1^\ast,A_2,A_3,R_4^\ast]\) with \(\beta=0.5\), \([R_1^\ast,R_2^\ast,A_3,R_4^\ast]\) with \(\beta=0.5\), \([R_1^\ast,R_2^\ast,R_3^\ast,R_4^\ast]\) with \(\beta=0.5\), \([R_1^\ast,A_2,A_3,R_4^\ast]\) with \(\beta=1.0\), \([R_1^\ast,R_2^\ast,A_3,R_4^\ast]\) with \(\beta=1.0\), and \([R_1^\ast,R_2^\ast,R_3^\ast,R_4^\ast]\) with \(\beta=1.0\).}
\label{fig:additional_prefix_ablation}
\end{figure}

\paragraph{Prefix composition variants visualization.} Figure~\ref{fig:additional_prefix_ablation} provides additional qualitative comparisons of the early-scale prefix configurations under two structural correction strengths, \(\beta=0.5\) and \(\beta=1.0\). Given the same content image \(I^A\) and style image \(I^B\), we compare \([R_1^\ast,A_2,A_3,R_4^\ast]\), \([R_1^\ast,R_2^\ast,A_3,R_4^\ast]\), and \([R_1^\ast,R_2^\ast,R_3^\ast,R_4^\ast]\). These examples further illustrate the content--appearance trade-off observed in the quantitative ablation: progressively rectifying more early scales strengthens appearance transfer, while increasing \(\beta\) places greater emphasis on the structural information from the content reference.

\begin{figure}[H]
\centering
\includegraphics[width=\textwidth]{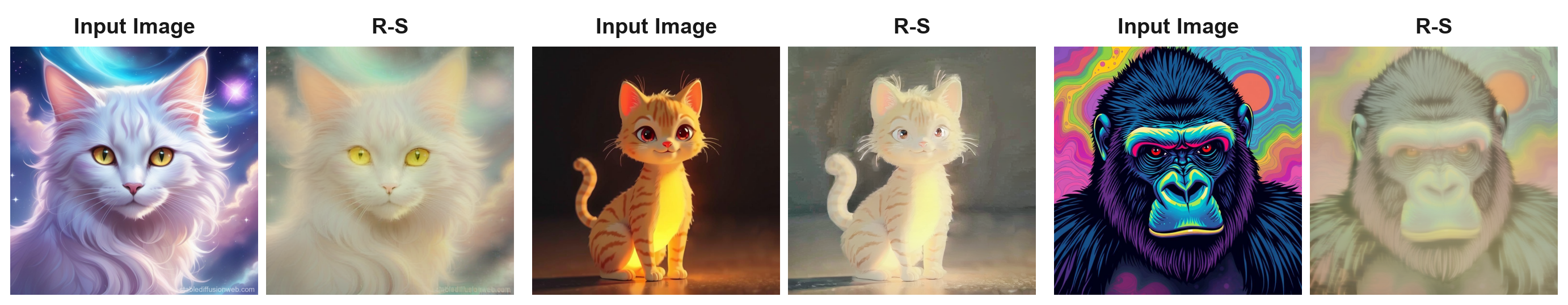}
\caption{\textbf{Visualization of style removal.} Each pair shows an input image and the corresponding style-removed result \(R-S\). The operation suppresses appearance-specific information while retaining the underlying object content and structure.}
\label{fig:style_removal}
\end{figure}

\paragraph{Style-removal visualization.} Figure~\ref{fig:style_removal} visualizes the effect of the estimated appearance component by comparing each input image with its style-removed representation \(R-S\). Removing \(S\) suppresses appearance-specific cues and moves the representation toward the neutral-appearance space, while retaining the dominant object structure. These examples provide a qualitative diagnostic of the appearance component estimated by our prompt-conditioned decomposition.

\section{Additional Qualitative Evaluation}
\label{app:qualitative_evaluation}

\begin{figure}[H]
\centering
\includegraphics[width=\textwidth]{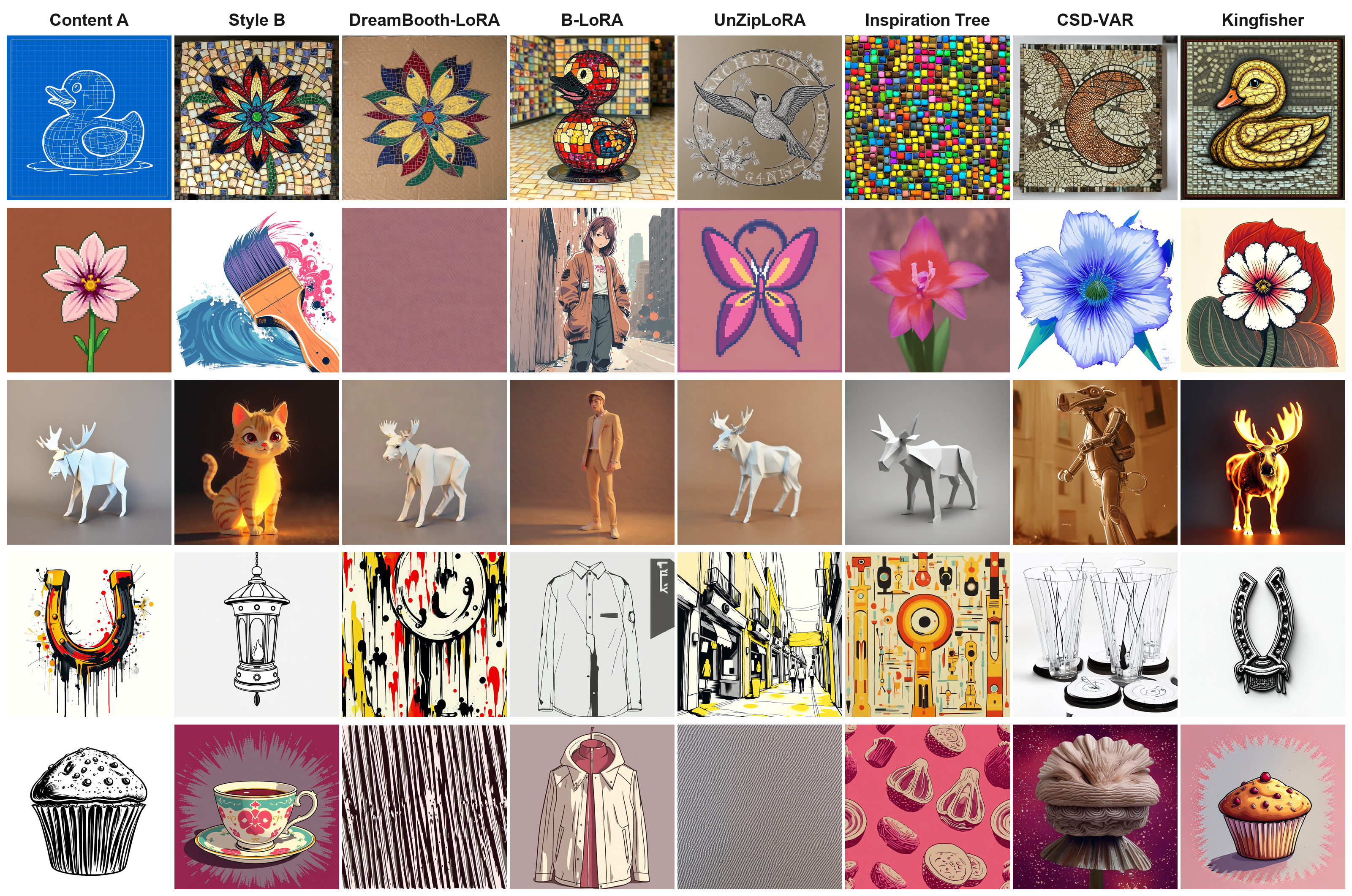}
\caption{\textbf{Additional qualitative comparison with prior methods.} Given a content reference \(I^A\) and a style reference \(I^B\), we compare Kingfisher with the evaluated baselines across additional content--style combinations.}
\label{fig:qualitative_evaluation}
\end{figure}

Figure~\ref{fig:qualitative_evaluation} presents additional qualitative comparisons with prior methods across diverse content--style pairs. The results further show that Kingfisher achieves a favorable balance between preserving the content structure of \(I^A\) and transferring the visual appearance of \(I^B\).

\section{Qualitative Results}
\label{app:additional_qualitative}

\begin{figure}[H]
\centering
\includegraphics[width=\textwidth]{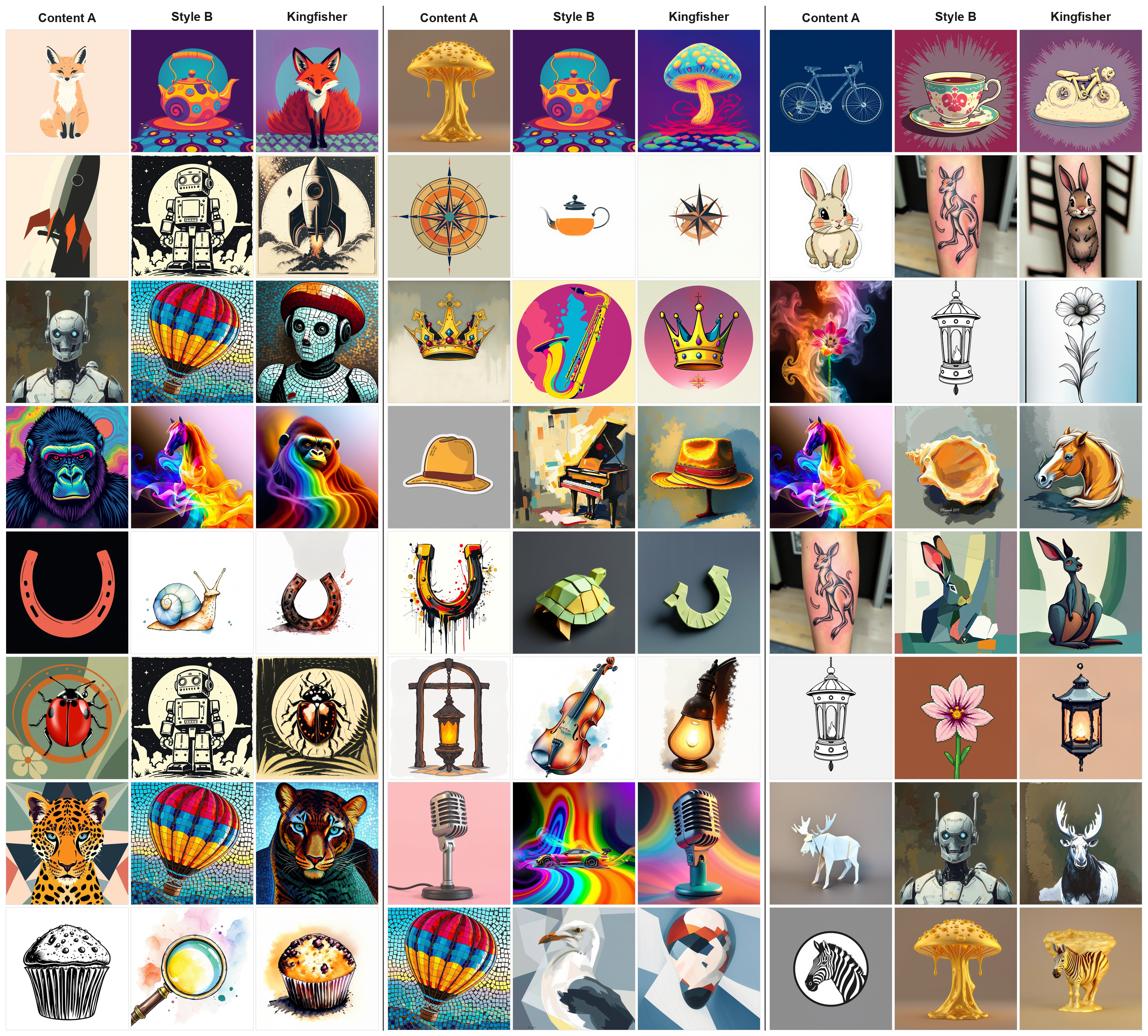}
\caption{\textbf{Qualitative results of Kingfisher on CSD-100.} Each example shows the content reference \(I^A\), the style reference \(I^B\), and the corresponding Kingfisher output.}
\label{fig:additional_qualitative}
\end{figure}

Figure~\ref{fig:additional_qualitative} presents qualitative results on the CSD-100 benchmark. Given a content reference \(I^A\) and a style reference \(I^B\), Kingfisher consistently preserves the main object structure of the content image while transferring the visual appearance of the style reference across diverse content--style combinations.

\section{Implementation Details}


For \textbf{Kingfisher}, Kingfisher is built on the publicly available pretrained \textbf{Infinity-2B} model~\cite{han2025infinity}, which is kept entirely frozen throughout all experiments. Infinity represents an image using \(N=13\) residual scales. Kingfisher performs prompt-conditioned decomposition and composition only on the early semantic prefix \(R_1\)--\(R_4\), while the remaining scales \(R_5,\ldots,R_{13}\) are autoregressively predicted by the pretrained Infinity-2B generator. We use five neutral prompts and six object prompts and retain the first four singular directions for both prompt-conditioned subspaces. The content and appearance components are optimized for 250 steps using Adam with a learning rate of \(3\times10^{-2}\). We set
\[
\lambda_{\mathrm{sty}}=1.0,\qquad \lambda_{\mathrm{align}}=0.5,\qquad \lambda_{\mathrm{remove}}=0.75,\qquad \lambda_{\mathrm{rec}}=0.05,\qquad \lambda_{\mathrm{orth}}=0.25.
\]
Our default composed prefix is
\begin{equation}
[R_1^\ast,R_2^\ast,A_3,R_4^\ast],
\end{equation}
where
\[
R_1^\ast=A_1-S_1^A+S_1^B,\qquad R_2^\ast=A_2-S_2^A+S_2^B,
\]
and
\[
R_4^\ast=B_4+\beta(C_4^A-C_4^B),\qquad \beta=0.5.
\]
Classifier-free guidance is set to \(3.0\). No model weights, adapters, K--V representations, or subject-specific embeddings are updated during personalization.

For \textbf{B-LoRA}, we optimize one rank-64 representation for each reference image for 1,000 steps with a learning rate of \(5\times10^{-5}\). Each image is associated with a generic learned identifier ``\texttt{A [vN]}''. For a pair \((I^A,I^B)\), we combine the official content block from \(I^A\) with the style block from \(I^B\), using ``\texttt{A [vA] in [vB] style}'' for generation.

For \textbf{DreamBooth-LoRA}, we learn role-specific content and style LoRA representations from each reference. The content representation is trained using ``\texttt{a photo of [cN]}'', while the appearance representation uses ``\texttt{an image in [sN] style}''. At inference, the content LoRA from \(I^A\) and the style LoRA from \(I^B\) are activated jointly, and the target is generated using ``\texttt{a photo of [cA] in [sB] style}''.

For \textbf{UnZipLoRA}, we use its SDXL implementation with rank 64 and 600 optimization steps per reference. Cross-image inference uses the content LoRA and merger learned from \(I^A\), together with the style LoRA and merger learned from \(I^B\).

The original \textbf{Inspiration Tree} implementation targets SD1.x; we therefore use an SDXL port following its original optimization principle. For each reference, four candidate trees are optimized for 200 steps, selected using image-based coherence and distinctness, and the selected initialization is replayed to 1,000 steps. During cross-image composition, the learned children from \(I^A\) and \(I^B\) are used as the content and style references, respectively.

For \textbf{CSD-VAR}, we report the official Switti backbone implementation. CSD-VAR decomposes each source image independently as
\begin{equation}
I^A \rightarrow (t_C^A,t_S^A), \qquad I^B \rightarrow (t_C^B,t_S^B),
\end{equation}
and generates the cross-image result using the content representation \(t_C^A\) and style representation \(t_S^B\). We retain the semantic caption and initializer conditioning used by the released CSD-VAR pipeline. 

\section{Limitations}
\begin{figure}[H]
\centering
\includegraphics[width=\textwidth]{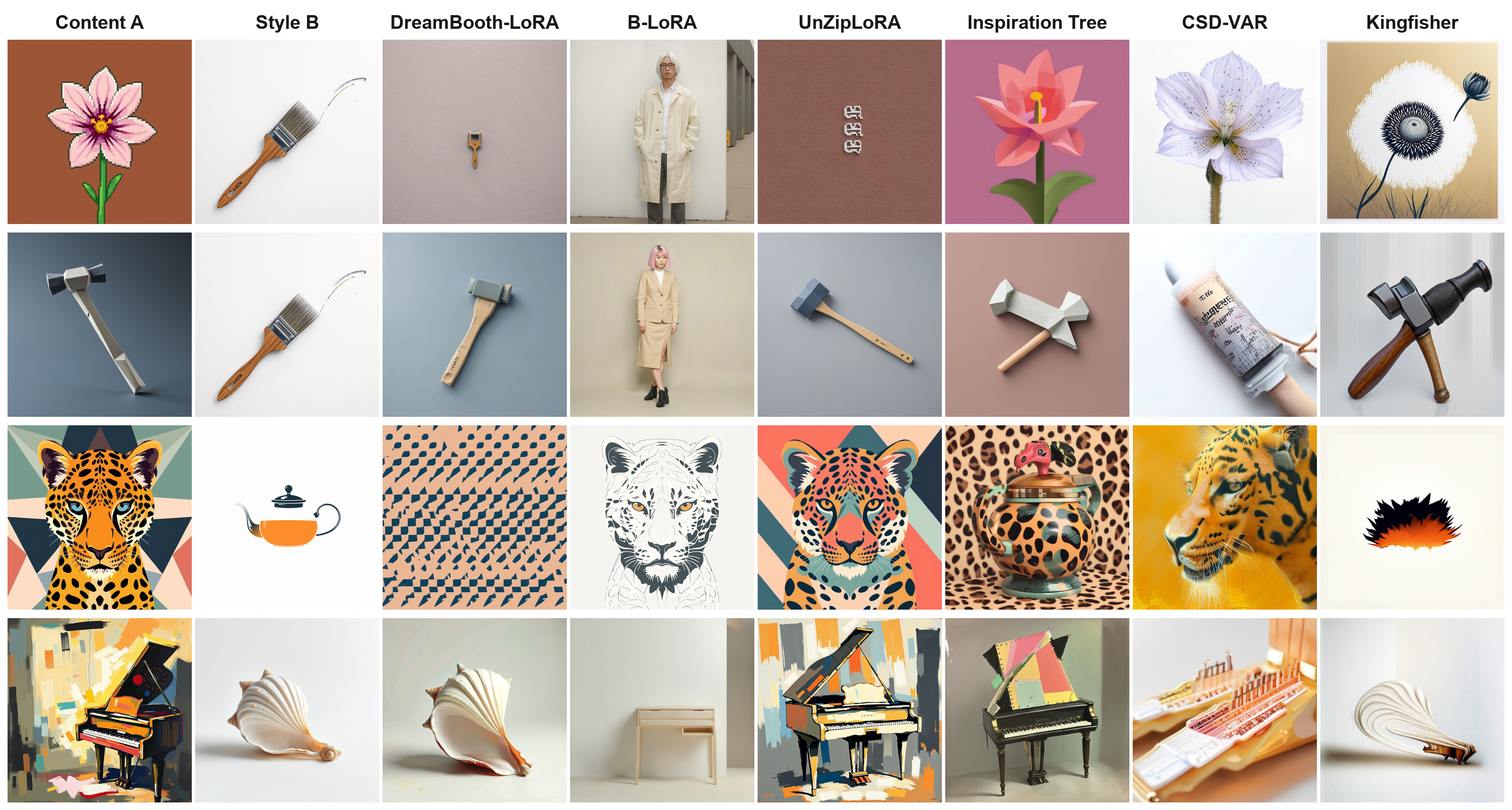}
\caption{\textbf{Challenging cases with abstract or ambiguous style references.} When the style reference contains weak, highly abstract, or insufficiently distinctive visual cues, its appearance is difficult to isolate from a single image. Consequently, both existing baselines and Kingfisher may exhibit weaker or less consistent style transfer.}
\label{fig:limitations}
\end{figure}

Kingfisher is less effective when the reference appearance is visually weak or insufficiently distinctive. In particular, styles with limited color variation, texture, illumination, or other characteristic appearance cues may induce only a small deviation from the neutral appearance space, making the estimated style component less discriminative and resulting in weaker perceptual transfer (as illustrated in Fig.~\ref{fig:limitations}). In addition, our decomposition relies on prompt-conditioned object directions; therefore, its effectiveness depends on how accurately the prompts describe the target content and how well the frozen generator responds to these descriptions.

\end{document}